\pdfoutput=1

\documentclass[11pt]{article}

\usepackage[]{acl}

\usepackage{times}
\usepackage{latexsym}
\usepackage{multirow}
\usepackage{graphicx}
\usepackage{enumitem}
\usepackage{amsmath}
\usepackage{amssymb}
\usepackage{latexsym}
\usepackage[inkscapelatex=false]{svg}
\usepackage{url}

\usepackage{ulem}   
\usepackage[linesnumbered,ruled,vlined]{algorithm2e} 
\usepackage{subfigure}
\usepackage{makecell}
\usepackage{color}
\usepackage{booktabs}
\usepackage{hyperref}
\usepackage{comment}
\usepackage[all]{nowidow}
\usepackage[subtle]{savetrees} 
\usepackage[T1]{fontenc}

\usepackage[utf8]{inputenc}

\usepackage{microtype}

\usepackage{inconsolata}

\SetKwInput{KwInput}{Input}
\SetKwInput{KwOutput}{Output}

\title{BioMNER: A  Dataset for Biomedical Method Entity Recognition}

\author{
 Chen Tang$^{1}$ , Bohao Yang$^{2}$, Kun Zhao$^{3}$, Bo Lv$^{4}$,  \textbf{Chenghao Xiao}$^{5}$, \\ \textbf{Frank Guerin}$^{1}$ \textbf{and} \textbf{Chenghua Lin}$^{2}$\thanks{Corresponding author.} \\
$^{1}$ University of Surrey, 
$^{2}$ The University of Manchester, $^{3}$ University of Pittsburgh,\\ $^{4}$ Chinese Academy of Sciences, $^{5}$ Durham University \\
\texttt{chen.tang@surrey.ac.uk}
}

\begin{document}

\maketitle
\begin{abstract}
Named entity recognition (NER) stands as a fundamental and pivotal task within the realm of Natural Language Processing. Particularly within the domain of Biomedical Method NER, this task presents notable challenges, stemming from the continual influx of domain-specific terminologies in scholarly literature. Current research in Biomedical Method (BioMethod) NER suffers from a scarcity of resources, primarily attributed to the intricate nature of methodological concepts, which necessitate a profound understanding for precise delineation. In this study, we propose a novel dataset for biomedical method entity recognition, employing an automated BioMethod entity recognition and information retrieval system to assist human annotation. Furthermore, we comprehensively explore a range of conventional and contemporary open-domain NER methodologies, including the utilization of cutting-edge large-scale language models (LLMs) customised to our dataset. Our empirical findings reveal that the large parameter counts of language models surprisingly inhibit the effective assimilation of entity extraction patterns pertaining to biomedical methods. Remarkably, the approach, leveraging the modestly sized ALBERT model (only 11MB), in conjunction with conditional random fields (CRF), achieves state-of-the-art (SOTA) performance.
\end{abstract}

\section{Introduction}
The task of extracting method entities from unstructured biomedical text represents a pivotal Natural Language Processing (NLP) challenge, warranting scholarly attention. Within the domain of biomedicine, Biomedical Method Named Entity Recognition (BioMethod NER) is of utmost importance, as the continuous influx of new terminology introduced through research publications makes it progressively more difficult for researchers to access pertinent information. This domain-specific task assumes a critical role in the automatic identification and extraction of methodological terminology, thereby enhancing the comprehensibility of biomedical literature to experts and laypeople alike.

In a notable study by \citet{wang2022review}, an in-depth exploration of method entities culminated in a clear definition, characterizing them as "named entities that represent specific methods". These method entities manifest as noun phrases denoting the specific techniques, approaches, mechanisms, and strategies employed in the resolution of tasks or problems proposed by authors. This encompasses sub-categories of discipline-specific methodologies, software tools, models, algorithms, and metrics, thus being commonplace in the literature \cite{kalim-mercer-2022-method}. Unlike open-domain NER, such as recognising the common entities of PERSON, LOCATION and ORGANISATION, the task of BioMethod NER faces additional challenges: (1) The language models are expected to overcome resource scarcity problems, as biomedical concepts are not often commonsense to the annotators~\cite{tang2023improving,goldsack2023enhancing}, leading to  time-consuming annotation work \cite{houngbo-mercer-2012-method}; (2) The increasing rate of newly created terminology requires that NER systems effectively learn patterns and rules to identify potential BioMethod NER entities \cite{tang2023terminology,yang2023improving}.

Currently, the field of BioMethod NER lacks comprehensive investigation. Prior research efforts~\cite{houngbo-mercer-2012-method,kalim-mercer-2022-method,wang2022review} have predominantly focused on conventional machine-learning-based NER approaches, such as Conditional Random Field (CRF)~\cite{Lafferty2001ConditionalRF}, and Bidirectional Long Short-Term Memory (BiLSTM)~\cite{graves2005framewise} methodologies. Additionally, to date, there has been a notable absence of a high-quality gold dataset for BioMethod NER, as highlighted by \citet{kalim-mercer-2022-method}\footnote{\citet{kalim-mercer-2022-method} claimed that 17 out of 20 predictions labelled as non-method entities in the gold-standard testing data of \citet{houngbo-mercer-2012-method} are actually method entities.}. In light of this, we endeavor to establish a high-quality dataset for the task of BioMethod NER. To guarantee the annotation quality of the bio-method entities, we propose a auxiliary annotation system which firstly use rules and ChatGPT~\footnote{An extremely large-scale language model available at \url{https://chat.openai.com/}. We use the version of ``gpt-3.5-turbo-0613''.} to identify potential bio-method entity candidates, and then retrieves their related information from ChatGPT and Wikipedia\footnote{Implemented via the python package \textit{wikipedia} \url{https://pypi.org/project/wikipedia/}.} for the reference of the annotators.
 Subsequently, we conduct experiments with a diverse array of conventional and contemporary methods, including Bi-LSTM~\cite{graves2005framewise}, CRF~\cite{lafferty2001conditional}, BERT~\cite{devlin-etal-2019-bert} and other large-scale language models (LLMs), to evaluate their abilities in tackling the challenges posed by the BioMethod NER task.

Our contributions are as follows:
\begin{itemize}
[noitemsep,topsep=0pt,parsep=0pt,partopsep=0pt,leftmargin=*]
\item We propose a novel auxiliary annotation system help annotators recognise the potential bio-method entities contained in biomeidcal literature, which bolster the annoting speed, and also increases the inter-agreement of the annotators.
\item We introduce a novel dataset with high-quality annotations, specifically designed for the BioMethod NER task. 
\item We employ a diverse range of machine learning techniques and conduct extensive experiments to analyze the performance of these approaches in addressing the challenges posed by the BioMethod NER task.
\end{itemize}

\section{Related Work}
Named Entity Recognition (NER) is a foundational task in Natural Language Processing (NLP), wherein entities are identified and labeled based on a mix of semantic and syntactic principles. NER has received a lot of research attention in the past three decades \cite{nasar2021named}. Nonetheless, in contrast to open-domain NER tasks, biomedical NER studies have received comparatively less scholarly attention, resulting in a disparity in methodological advancement \cite{song2021deep}. Therefore, this paper endeavors to build upon recent developments in deep learning within open-domain NER, aiming to ascertain their efficacy in the context of BioMethod NER. 

NER is traditionally approached as a sequence labeling task, where language models featuring encoder architectures such as BERT~\cite{devlin-etal-2019-bert}, ALBERT~\cite{lan2019albert}, and RoBERTa~\cite{liu2019roberta} are  employed within the token classification module. Additionally, we acknowledge the inclusion of conventional methodologies, including Long Short-Term Memory (LSTM) networks~\cite{graves2005framewise} and Conditional Random Fields (CRF)~\cite{lafferty2001conditional}, as their deep learning counterparts have challenges in handling few-shot scenarios \cite{tang-etal-2022-ngep,loakman-etal-2023-twistlist}.  Furthermore, we recognise the emergence of exceptionally large-scale language models, exemplified by ChatGPT, boasting a staggering 100 billion (100B) parameter count. These models have introduced remarkable advancements in natural language generation, with recent investigations~\cite{wang2023gpt, zhou2023universalner} exploring the utilization of ChatGPT for NER as a generative task. Consequently, our research encompasses the evaluation of ChatGPT's zero-shot capabilities, alongside the fine-tuning of a 7-billion-parameter language model, Baichuan~\cite{yang2023baichuan}, to conduct our experiments.


\begin{figure}[t]
\centering
\includegraphics[width=0.99\columnwidth]{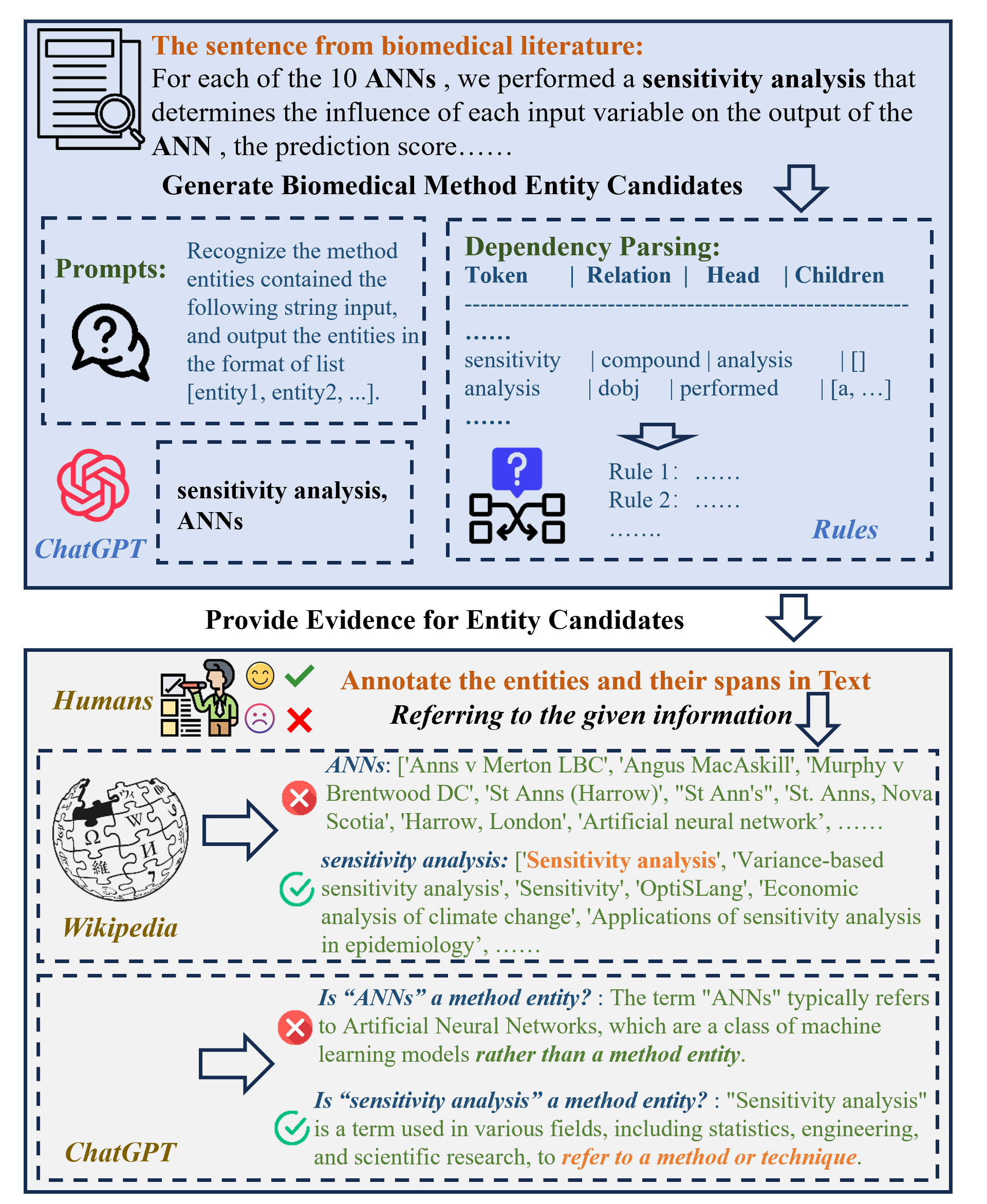}
\caption{An illustration of our proposed automatic biomedical entity NER system.}
\label{fig:data}
\end{figure}

\section{Dataset Construction}
The raw biomedical text corpus is sourced from a set of sentences collected by \citet{kalim-mercer-2022-method}, which potentially contain biomedical entities. Directly annotating the method entities from biomedical literature is a challenging task for human annotators, demanding not only an extensive theoretical understanding of biomedicine but also a proficiency in the specialised field. In order to expedite the annotation workflow while maintaining data quality, we introduce a systematic framework illustrated in \autoref{fig:data}, which leverages the capabilities of LLMs, i.e. ChatGPT, in conjunction with rule-based\footnote{We follow \citet{kalim-mercer-2022-method}, where biomedical methods are extracted using linguistic rules and patterns during dependency parsing.} unsupervised methods to identify potential candidates from the sentences extracted from selected biomedical literature. Subsequently, the identified biomedical method entity candidates undergo a validation process by consulting ChatGPT and Wikipedia to retrieve relevant information in the biomedical domain, thereby distinguishing them from other non-relevant entity phrases or abbreviations. Following this, the recognized entity candidates and the retrieved information are validated by humans to complete the annotation process. In a nutshell, this validation process relies on human judgment based on cross-referencing results obtained from Wikipedia and explanations provided by ChatGPT. Three annotators participated in the data annotation process, and only instances with consistent annotations from all evaluators were retained for inclusion in the dataset. Merely 64 samples were excluded from the raw corpus, accounting for a proportion of 0.0482. The statistical overview of this dataset is presented in \autoref{tab:data_stat}. 

To illustrate the data construction pipeline, we present a representative example shown in \autoref{fig:data}. The input text for this example is as follows: ``For each of the 10 ANNs, we performed a sensitivity analysis that determines the influence of each input variable on the output of the ANN, the prediction score.'', The first step is to identify biomedical entity candidates, with the rule-based method identifying ``sensitivity analysis'' as a potential candidate, and ChatGPT providing ``ANNs'' and "sensitivity analysis" as the candidates. In the second step, we retrieve the information of these entity candidates. When consulting Wikipedia, the search results of ``ANNs'' are ``Anns v Merton LBC'' and ``Angus MacAskill'', which has no relation to a method, but the entry of ``sensitivity analysis'' exists. Following this, we let ChatGPT  discern if these are method candidates specifically, and we get a rejection for ``ANNs'', and a confirmation for ``sensitivity analysis''. Consequently, during the human annotation stage, after manually reviewing the NER candidates and their relevant information provided by Wikipedia and ChatGPT, "sensitivity analysis" is confirmed as the ground truth result.

\begin{table}[tb]
\scriptsize \centering
\resizebox{0.99\linewidth}{!}{
\begin{tabular}{l|cccc}
\toprule
\textbf{Datasets}&\textbf{Train} & \textbf{Test} & \textbf{All}\\
\midrule
\textbf{\# Samples}    & 1128 & 134 & 1262 \\
\textbf{\# Vocabulary} & 5743 & 1449 & 6197  \\
\textbf{\# Entities}   & 1010 & 130 & 1119  \\
\midrule
\textbf{Avg. \# Words in A Sample}     & 25.75 & 27.51 & 25.94   \\
\textbf{Avg. \# Entity Words in A Sample} & 2.89 & 2.79 & 2.88    \\
\textbf{Avg. \# Entity Proportion in A Sample} & 0.1122 & 0.1014 & 0.1110    \\
\bottomrule
\end{tabular}
}
\caption{Data Statistics of the annotated dataset. Avg. is the abbreviation of Average. Entity Proportion represents the ratio of entity words in the total works in a sample.}
\label{tab:data_stat}
\end{table}

\section{Methodology}
The methodologies employed in BioMethod NER can be categorised into two primary groups as follows: \textbf{(1) Sequence Labeling:} This group encompasses a range of methods, including \textbf{Bi-LSTM}\footnote{Bi-LSTM is the abbreviation of Bi-directional Long Short Term Memory network. It is a widely used recurrent neural network in the sequence labelling task.}~\cite{graves2005framewise}, \textbf{CRF}~\cite{lafferty2001conditional}, \textbf{BERT}~\cite{devlin-etal-2019-bert}, \textbf{BioBERT}~\cite{lee2020biobert}, \textbf{ALBERT}\footnote{Albert has two version with different experimental settings. We select the latest version (version) settings here}~\cite{lan2019albert}, and \textbf{RoBERTa}~\cite{liu2019roberta}, as well as variations that combine CRF with BERT, ALBERT, and RoBERTa. \textbf{(2) Text Generation:}  In this category, we leverage advanced models such as \textbf{ChatGPT} in a zero-shot setting and the latest version (version 2) of Baichuan~\cite{yang2023baichuan}.\footnote{\url{https://github.com/baichuan-inc/Baichuan2}}

Within the framework of Sequence Labeling models, the entity list target adheres to the ``BIO'' tagging format, where each token is assigned a label based on the ``BIO'' tagging scheme\footnote{Specifically, tokens are categorised as follows: Tokens at the start of a named entity are denoted as ``B-method''.
Tokens residing inside a named entity but not at the entity's onset are assigned the label ``I-method''. Tokens not associated with any named entity are simply labeled as ``O''.}~\cite{lample-etal-2016-neural}.  In the context of text generation models, the target format for the entity list adopts a distinct structure. Instead of employing BIO tagging, entities are presented as a concatenated string, organized as "entity 1, entity 2, ...". This format is optimal for text generation tasks, where the goal is to generate coherent and human-readable text containing the specified entities.

\begin{figure}[t]
\centering
\includegraphics[width=0.7\columnwidth]{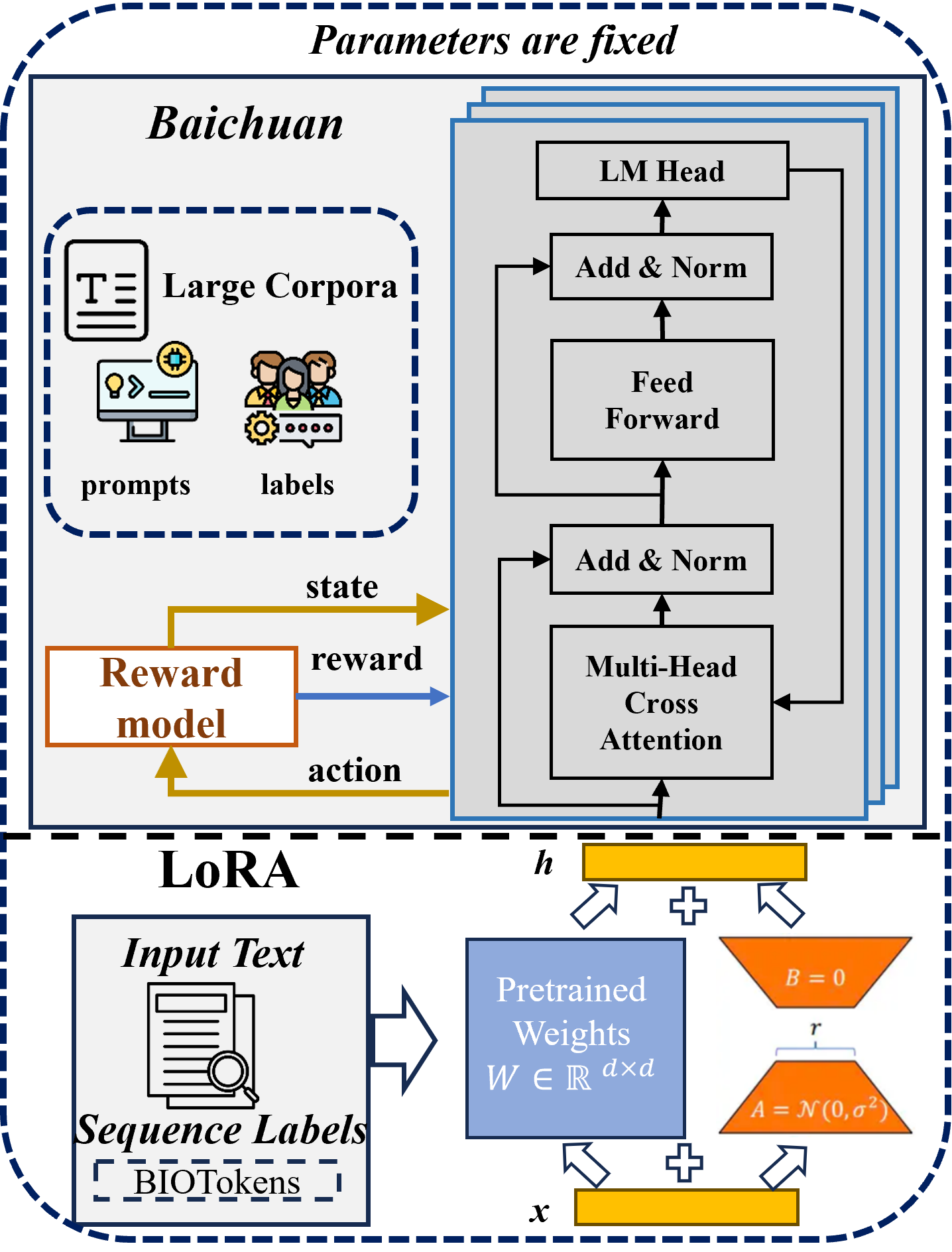}
\caption{An illustration of the Baichuan+LoRA method.}
\label{fig:baichuan}
\end{figure}

In light of the considerable computational resources demanded by fine-tuning  the 7B Baichuan model, we have opted for an alternative approach shown in \autoref{fig:baichuan} by incorporating the LoRA technique~\cite{hu2021lora}. The LoRA technique involves the immobilization of pre-trained model weights, coupled with the introduction of trainable rank decomposition matrices within each layer of the Transformer architecture. This strategic implementation effectively curtails the volume of trainable parameters, thereby rendering it a more resource-efficient solution for addressing downstream tasks. More details of the methodologies refer to Appendix \ref{apx:method}.

\section{Experimental Results}

\begin{table}
\centering
\resizebox{0.7\linewidth}{!}{
\begin{tabular}{r|ccc|c}
\toprule[1pt]
\multirow{2}{*}{\textbf{Systems}} & \multicolumn{4}{c}{\textbf{Metrics}}  \\
\cline{2-5} 
& \textbf{Precision} & \textbf{Recall} & \textbf{F1} & \textbf{Acc}   \\
\midrule
\textbf{ChatGPT$_{zero}$}  & 34.09 & 5.13 & 8.91 & 72.63   \\
\textbf{Baichuan+LoRA}  & 68.25 & 65.15 & 66.67 & 94.52 \\
\midrule
\textbf{CRF}  &  81.75  & 75.81  & 78.31 & 95.21   \\
\textbf{Bi-LSTM+CRF}  & 77.14  &  83.57  &  80.23 & 96.76 \\ 
\textbf{DistilBERT}  & 76.62 & 79.72 & 76.98 & 97.16 \\
\textbf{BERT}  & 76.44 & 78.94 & 77.19 & 97.40 \\ 
\textbf{Bio-BERT}  & 82.46 & 85.10 & 82.74 & 98.25 \\ 
\textbf{RoBERTa}  & 80.50 & 83.06 & 81.33 & 98.05 \\
\textbf{AlBERT}  & 92.40 & 95.95 & 93.76 & 99.21 \\
\textbf{DistilBERT+CRF} & 75.90 & 77.66 & 76.03 & 97.12  \\
\textbf{BERT+CRF} & 78.54 & 82.77 & 80.03 & 97.57     \\ 
\textbf{Bio-BERT+CRF} & 84.60 & 86.55 & 85.22 & 98.34 \\ 
\textbf{RoBERTta+CRF}  & 83.05 & 85.61 & 84.06 & 98.11   \\
\textbf{AlBERT+CRF}  &  \textbf{94.51} & \textbf{96.72} & \textbf{95.36} & \textbf{99.36}  \\
\bottomrule[1pt]
\end{tabular}
}
\caption{Automatic evaluation results. \textbf{F1} denotes the F1 score, and \textbf{Acc} denotes the accuracy. The best performing solutions are in \textbf{bold.}}
\label{tab:auto_eval}
\end{table}

The experimental results of all baseline models are presented in \autoref{tab:auto_eval}. It is noteworthy that the performance of Large Language Models (LLMs), specifically ChatGPT and Baichuan with LoRA, is inferior to that of the sequence labeling models. Despite ChatGPT being a powerful language model with 130 billion parameters, its performance in recognizing biomedical method entities is notably subpar, with a lack of data yielding accurate results. In contrast, we also assessed ChatGPT's performance on an open-domain dataset (the CoNLL 2023 NER dataset~\cite{tjong-kim-sang-de-meulder-2003-introduction}), where it demonstrated exceptional proficiency in identifying PERSON, LOCATION, and ORGANIZATION entities without requiring additional training, a phenomenon corroborated by recent studies~\cite{wang2023gpt, zhou2023universalner}.

Similarly, Baichuan's performance does not align with that of its sequence labeling counterparts. We posit that the success of LLMs is primarily attributed to the vast knowledge they acquire during extensive pretraining. However, these models grapple with adapting their reasoning capabilities to domain-specific NER tasks due to the inherent constraints of their pretrained parameters, which impedes their capacity to efficiently learn and apply domain-specific knowledge patterns. This assumption finds further validation when scrutinizing traditional sequence labeling tasks. For instance, Conditional Random Fields (CRF), a statistical approach that leverages rule-based features to discern patterns, outperforms several powerful representation-based language models such as BERT and DistilBERT. This outcome is underpinned by the high precision score of CRFs, indicating that entities recognized by CRFs are more likely to align with the ground truth.

Moreover, all language models, when combined with CRF, exhibit significant improvements across all tested metrics. This underscores the efficacy of explicitly modeling entity patterns (e.g., CRF) in contrast to the implicit pattern modeling facilitated by pretraining (e.g., BERT). Comparing the language models directly, it becomes apparent that, with or without CRF, their ranking (based on F1 scores) is as follows: DistilBERT, BERT, Bi-LSTM, RoBERTa, Bio-BERT and ALBERT. Overall, ALBERT attains state-of-the-art performance on our dataset.

\section{Conclusion}

In conclusion, we present a novel dataset tailored for the Biomedical Method Named Entity Recognition (BioMethod NER) task, alongside the introduction of an auxiliary annotation system to facilitate human annotation. We conduct the experiments with a range of traditional and contemporary open-domain NER models, including the utilization of large-scale language models (LLMs) customized to our dataset. Notably, our empirical findings show the hindrance posed by the excessive parameter size of large-scale language models on the effectiveness of entity extraction patterns pertaining to biomedical methods.

\normalem
\bibliography{custom}

\clearpage

\appendix
\section{Appendices}
\label{sec:appendix}

\subsection{Methodologies} \label{apx:method}

Recent advancements in language models\cite{yang2023effective,tang-etal-2022-ngep,huang-etal-2022-improving,tang-etal-2022-etrica,tang2024cross,tang-etal-2023-enhancing,zhao2024slide,yang2024structured,loakman2024train} primarily rely on extensive pre-training processes to imbue these models with fundamental language comprehension and generation capabilities. To ensure a consistent baseline performance, all of the language models (LMs) utilised in our experiments were sourced from publicly available checkpoints on Hugging Face.\footnote{\url{https://huggingface.co/models}}
These models underwent training over a span of up to 30 epochs, conducted on a Tesla A40 machine (with the exception of Baichuan, which was trained on a Tesla A100 machine), with 48 GB of GPU memory. The batch size was configured at 16, while the learning rate was set to $1e^{-4}$, with training executed with the Adam optimizer. Additionally, a linear learning rate scheduler was employed, with no warm up steps. The specific checkpoints selected for our study are as follows: 

\begin{itemize}[noitemsep,nolistsep,leftmargin=*]
\item \textbf{DistilBERT/+CRF: } \url{https://huggingface.co/distilbert-base-uncased};
\item \textbf{BERT/+CRF: } \url{https://huggingface.co/bert-base-uncased}; 
\item \textbf{Bio-BERT/+CRF: } \url{https://huggingface.co/dmis-lab/biobert-base-cased-v1.2}; 
\item \textbf{ALBERT/+CRF: } \url{https://huggingface.co/albert-base-v2};  
\item \textbf{RoBERTa/+CRF: } \url{https://huggingface.co/roberta-base}; 
\item \textbf{Baichuan: } \url{https://huggingface.co/baichuan-inc/Baichuan-7B};
\end{itemize}

These selected checkpoints, coupled with our chosen methodologies, form the basis of our experimental framework, which aims to provide valuable insights into the task of BioMethod NER.

\end{document}